\documentclass[conference]{IEEEtran}

\usepackage{amsmath,amsfonts,bm}

\def\eqref#1{equation~\ref{#1}}

\def\1{\bm{1}}

\def\vmu{{\bm{\mu}}}

\def\vb{{\bm{b}}}

\def\vt{{\bm{t}}}

\def\vx{{\bm{x}}}

\def\vz{{\bm{z}}}

\def\mA{{\bm{A}}}

\def\mP{{\bm{P}}}

\def\mT{{\bm{T}}}

\def\mW{{\bm{W}}}
\def\mX{{\bm{X}}}

\def\mSigma{{\bm{\Sigma}}}

\DeclareMathAlphabet{\mathsfit}{\encodingdefault}{\sfdefault}{m}{sl}
\SetMathAlphabet{\mathsfit}{bold}{\encodingdefault}{\sfdefault}{bx}{n}
\newcommand{\tens}[1]{\bm{\mathsfit{#1}}}
\def\tA{{\tens{A}}}

\usepackage{graphicx}
\usepackage{amsmath}
\usepackage{amssymb}
\usepackage{booktabs}

\usepackage[pagebackref,breaklinks,colorlinks]{hyperref}

\usepackage[capitalize]{cleveref}
\crefname{section}{Sec.}{Secs.}
\Crefname{section}{Section}{Sections}
\Crefname{table}{Table}{Tables}
\crefname{table}{Tab.}{Tabs.}
\crefname{appendix}{App.}{Apps.}

\newcommand{\enquote}[1]{"#1"}

\newcommand{\vgamma}{{\bm{\gamma}}}

\newcommand{\vpi}{{\bm{\pi}}}

\newcommand{\circled}[1] {\raisebox{.5pt}{\textcircled{\raisebox{-.3pt} {\scriptsize #1}}}}

\let\oldsqrt\sqrt
\def\sqrt{\mathpalette\DHLhksqrt}
\def\DHLhksqrt#1#2{%
	\setbox0=\hbox{$#1\oldsqrt{#2\,}$}\dimen0=\ht0
	\advance\dimen0-0.2\ht0
	\setbox2=\hbox{\vrule height\ht0 depth -\dimen0}%
	{\box0\lower0.4pt\box2}}

\begin{document}

\title{A new perspective on probabilistic image modeling}

\author{\IEEEauthorblockN{Alexander Gepperth}
\IEEEauthorblockA{\textit{Applied Computer Science Department} \\
\textit{University of Applied Sciences Fulda}\\
Fulda, Germany \\
alexander.gepperth@cs.hs-fulda.de}
}
\maketitle

\begin{abstract}
We present the Deep Convolutional Gaussian Mixture Model (DCGMM), a new probabilistic approach for image modeling capable of density estimation, sampling and tractable inference. DCGMM instances exhibit a CNN-like layered structure, in which the principal building  blocks are convolutional Gaussian Mixture (cGMM) layers. A key innovation w.r.t. related models like sum-product networks (SPNs) and probabilistic circuits (PCs) is that each cGMM layer optimizes an independent loss function and therefore has an independent probabilistic interpretation. This modular approach permits intervening transformation layers to harness the full spectrum of 
(potentially non-invertible) mappings available to CNNs, e.g., max-pooling or half-convolutions. DCGMM sampling and inference are realized by a deep chain of hierarchical priors, where a sample generated by a given cGMM layer defines the parameters of sampling in the next-lower cGMM layer. For sampling through non-invertible transformation layers, we introduce a new gradient-based sharpening technique that exploits redundancy (overlap) in, e.g., half-convolutions.
DCGMMs can be trained end-to-end by SGD from random initial conditions, much like CNNs. We show that DCGMMs compare favorably to several recent PC and SPN models in terms of inference, classification and sampling, the latter particularly for challenging datasets such as SVHN. We provide a public TF2 implementation.
\end{abstract}


\section{Introduction}
\noindent This conceptual work is in the context of probabilistic image modeling by a hierarchical extension of Gaussian Mixture Models (GMMs), which we term Deep Convolutional Gaussian Mixture Model (DCGMM). Main objectives are density estimation, image generation (sampling) and tractable inference (e.g., image in-painting). 
\par
Recent approaches (e.g., GANs, VAEs) excel in image generation by harnessing the full spectrum of CNN transformations, such as convolution or pooling. 
An issue is however the lack of density estimation and tractable inference capacity, i.e., explicitly expressing and exploiting the learned probability-under-the-model $p(\vx)$ of an image $\vx$. 
In contrast, recent \enquote{deep} probabilistic approaches like sum-product-networks and probabilistic circuits \cite{peharz2020random,poon2011sum,wolfshaar2020deep,peharz2020einsum,butz2019deep} have been explicitly designed to perform these functions on images. However, strong constraints must be satisfied at every hierarchy level to maintain a global probabilistic interpretation, which
excludes, e.g.,  overlapping convolutions and pooling.
We aim at overcoming these limitations by constructing deep hierarchies of convolutional Gaussian Mixture Models (GMMs) in order to perform density estimation, realistic sampling and inference within a single model for large-scale, complex visual problems. 
%
\subsection{DCGMM: Model overview and salient points}\label{sec:intro_salient}
\begin{figure}
\centering
\includegraphics[page=1,viewport=0in 6.7in 7.8in 8.2in, width=0.5\textwidth]{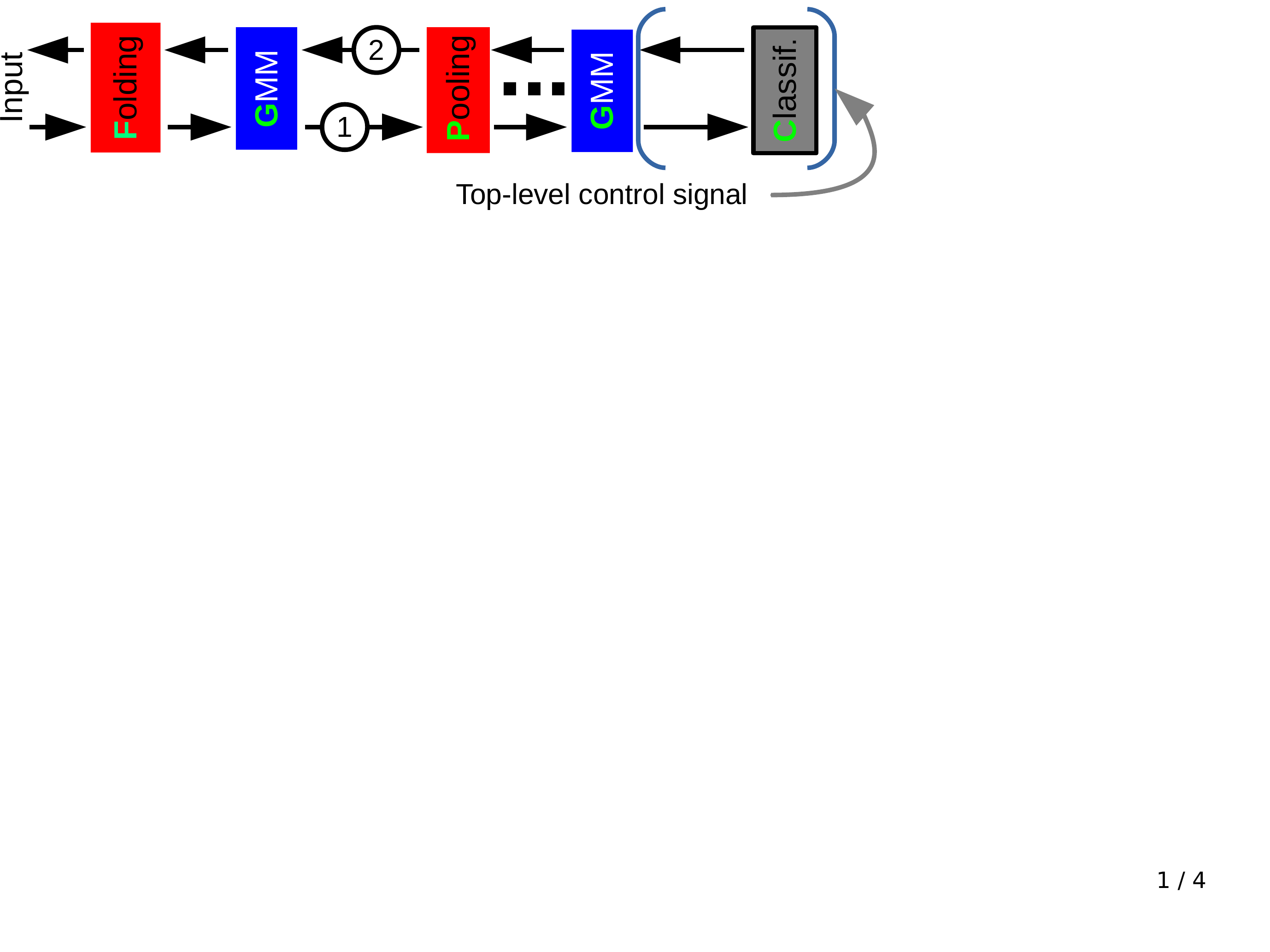}
\caption{\label{fig:architecture}
High-level structure of a typical DCGMM instance with principal layer types. DCGMMs have two operation modes: \circled{1}the forward mode which estimates the probability of a presented data sample \circled{2}the backwards mode in which a top-level control signal is propagated in reverse direction to generate a sample according to the learned model. The whole architecture is trained end-to-end, with each trainable layer being independently optimized.
}
\end{figure}
DCGMM instances can operate in density estimation (\enquote{forward}) and sampling (\enquote{backwards}) mode, and are realized by a succession of convolutional Gaussian Mixture Model (cGMM) and transformation layers, see \cref{fig:architecture}. 
\smallskip
\par\noindent Most prominently, DCGMMs optimize \textbf{independent loss functions} (no back-propagation) for all cGMM layers. 
This ensures a probabilistic interpretation of cGMM layer outputs and removes the need for structural constraints as imposed in
related models (see \cref{sec:relwork}).
\smallskip
\par\noindent Since DCGMMs relax the assumption of a single loss function, they can perform \textbf{standard CNN operations} between cGMM layers, e.g., half-convolution and pooling, which are non-invertible to different degrees.
\smallskip
\par\noindent In order to allow \textbf{sampling through non-invertible operations}, we exploit the fact that DCGMM convolutions are unconstrained and can thus be overlapping as in CNNs. We introduce a novel gradient ascent technique that exploits the redundancy thus created to compensate for the information loss due to non-invertible transformations.
%
\smallskip
\par
\begin{figure}[b]
\centering
\includegraphics*[page=4,viewport=0in 5.8in 7.8in 8.25in, width=0.5\textwidth]{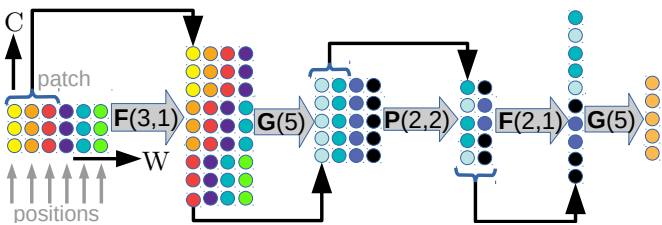}
\caption{\label{fig:convGMM}
A simple DCGMM instance applied to a NHWC tensor with N and W dimensions omitted. It contains max-pooling(P), half-convolution/folding(F) and cGMM (G) layers, see text for details.
Lower cGMM layers analyze local image \textit{patches} at different \textit{positions}. The topmost cGMM layer is global in the sense that it has a single output position only, in which  the whole image is described. 
Each cGMM layer provides a \enquote{fresh} probabilistic interpretation, indicated by uniform coloring for each position.
}
\end{figure}

\noindent A cGMM layer implements \textbf{parameter sharing}, using the same $\pi_k$, $\mSigma_k$ and $\vmu_k$ for all \textit{positions} of a four-dimensional NHWC input tensor (see \cref{fig:convGMM}).
This allows powerful models to be trained with few adjustable parameters.
%
%
\smallskip
\par\noindent DCGMM sampling realizes a \textbf{deep chain of hierarchical priors}:
A cGMM layer propagates posterior probabilities, for which the next upstream cGMM learns a model. A sample drawn from the upstream cGMM thus represents a likely realization of these posteriors. As such, it is a natural choice for controlling sampling in the current layer. 
%
%
\subsection{Related Work}\label{sec:relwork}

\noindent\textbf{GANs and VAEs and related models} The currently most widely used models of image generation are GANs and VAEs \cite{arjovsky2017wasserstein,Mirza2014,Goodfellow2014,kingma2014autoencoding}. 
GANs can sample photo-realistic images \cite{Richardson2018}, but are incapable of density estimation. 
VAEs show similar performance when it comes to sampling, 
and outlier detection can be performed, although general density estimation with VAEs is problematic as well . 
Inference with VAEs is complicated and potentially inefficient. 
Approaches with similar strength and weaknesses are realized by the PixelCNN architecture\cite{oord2016conditional}, GLOW \cite{kingma2018glow} and variants.
\smallskip
%
%
\smallskip\par\noindent\textbf{Hierarchical MFA/GMM}
Two related approaches are described in \cite{Viroli2019,VanDenOord2014}.  
These approaches realize hierarchical MFA as modular decompositions of single \enquote {flat} MFAs, and thus possess a single loss function that is optimized by Expectation-Maximization (EM). 
Closer to our approach is the proposal of \cite{Tang2012}, which proposes to train an MFA layer on the inferred latent variables of another, independent MFA instance. 
Transformations occurring in these models are global, i.e., affect \textit{all} input variables. Local operations like convolutions or max-pooling are not used, and the dimensionality of samples treated in the experiments is low.
A preliminary version of DCGMMs was described in \cite{gepperth2021a}.

\smallskip\par
\noindent\textbf{Probabilistic circuits and sum-product networks}
Probabilistic circuits are deep directed graphs whose leaves compute probabilities from inputs via tractable parameterized distributions. 
These probabilities are further processed at weighted-sum- and product nodes, with the aim of obtaining a tractable
distribution at the root node. This is ensured if some structural constraints are met, and the resulting PCs are often termed Sum-Product Networks
(SPNs, see, \cite{poon2011sum}). In particular, SPNs allow efficient sampling, density estimation and tractable inference even if graphs are very deep. 

Similarly to neural networks, finding a graph structure suitable for a particular learning task is challenging, especially if data dimensionality is high. 
Several interesting ideas for this have been put forward in \cite{peharz2020random,poon2011sum,wolfshaar2020deep,peharz2020einsum,butz2019deep}.
An appealing approach is realized by RAT-SPNs \cite{peharz2020random} which construct random SPNs that are then combined by a single root node, at the price of additional hyper-parameters.
Efficient learning and inference in SPNs is addressed by Einsum Networks \cite{peharz2020einsum} where successive sum and product nodes are combined into a GPU-friendly \textit{Einsum operation}.

SPNs suffer from the structural constraints required for representing valid, tractable distributions. These exclude overlapping convolutions and max-pooling.
Convolutions can be performed in special cases \cite{butz2019deep,wolfshaar2020deep}, but general CNN-like convolutions remain inaccessible.

SPNs have been successfully trained on high-dimensional visual problems \cite{peharz2020random,wolfshaar2020deep,peharz2020random,butz2019deep} like the MNIST and FashionMNIST benchmarks. 
Sampling and inference in SPNs are mostly demonstrated for the \enquote{Olivetti faces} benchmark \cite{butz2019deep,wolfshaar2020deep,poon2011sum} which is high-dimensional but contains only a few hundred samples.
To the best of our knowledge, sampling has not been demonstrated for MNIST and FashionMNIST.
In \cite{peharz2020einsum}, sampling and inference is demonstrated for the SVHN benchmark using a simple SPN and extensive data pre-processing. Generated samples 
are of good quality but not yet comparable to GANs or VAEs.
%
An interesting overview of the research in hierarchical generative mixture models is given in \cite{jaini2018deep}. 

\subsection{Objective, Contribution and Novelty}
\noindent The objectives of this article are to introduce a deep GMM architecture which exploits the same principles that led to the performance explosion of CNNs. 
Novel points are:
\begin{itemize}
\setlength\itemsep{0em}
  \item fundamentally new approach to hierarchical mixture modeling based on independent losses
  \item fundamentally new approach to sampling as a deep chain of hierarchical priors
	\item use of arbitrary transformations like convolution and pooling in probabilistic image modeling
	\item realistic sampling for complex visual tasks like SVHN
\end{itemize}

\section{Datasets}
\noindent For the evaluation we use the following image datasets:
\smallskip\\
\noindent\textbf{MNIST}~\cite{LeCun1998} consists of $60\,000$ $28$\,$\times$\,$28$ gray scale images of handwritten digits (0-9).
\textbf{FashionMNIST}~\cite{Xiao2017} consists of images of clothes in 10 categories and is structured like the MNIST dataset. 
\textbf{SVHN}~\cite{Netzer2011} contains color images of house numbers ($0$-$9$, resolution $32$\,$\times$\,$32$).
\smallskip\\
These datasets are not particularly challenging w.r.t. classification, but their dimensionality of $784$ (MNIST, FashionMNIST) and 3072 (SVHN)  is high, and the variability of SVHN, in particular, is considerable.
\section{Methods: DCGMM}
In order to introduce DCGMMs as a possible hierarchical generalization of GMMs, we will first review facts about vanilla GMMs and introduce the basic notation, 
which will subsequently be generalized to a multi-layered structure.
\subsection{Review of GMMs}\label{sec:intro:gmm}
GMMs aim to explain the observed data $X$\,$=$\,$\{\vx_n\}$ by expressing their density as a weighted mixture of $K$ Gaussian component densities $\mathcal{N}(\vx_n;\vmu_k,\mSigma_k)$\,$\equiv$\,$\mathcal{N}_k(\vx_n)$: 
\begin{equation}\label{eqn:gmm}
p(\vx_n)=\sum_{k=1}^K \pi_k \mathcal{N}_k(\vx_n),
\end{equation}
where the normalized component weights $\pi_k$ modulate the overall influence of each component.

GMMs assume that each observed data sample $\{\vx_n\}$ is drawn from one of the $K$ Gaussian component densities $\mathcal N_k$.
The selection of this component density is assumed to depend on an unobservable \textit{latent variable} $z_n\in\{1,\dots, K\}$ which follows an unknown distribution.
The \textit{complete-data likelihood} for a single data sample reads:
\begin{align}\label{eqn:complete}
p(\vx_n,z_n) &= \pi_{z_n}\mathcal{N}_{z_n}(\vx_n)  \\
p(\mX,\vz) &= \prod_n p(\vx_n,z_n)
\end{align}
For a single sample, Bayes' theorem gives us the \textit{GMM posteriors} or \textit{responsibilities} $\vgamma(\vx_n){=}p(z_n{=}k|\vx_n)$:
\begin{align}
\gamma_k(\vx_n)= \frac{p(\vx_n,z_n{=}k)}{\sum_j p(\vx_n,z_n{=}j)}{=}\frac{\pi_k \mathcal N_k(\vx_n)}{\sum_j \pi_j \mathcal N_j(\vx_n)}, 
\end{align}
which can be computed without the latent variables.
Marginalizing the unobservable latent variables out of \cref{eqn:complete}, we obtain the \textit{incomplete-data log-likelihood} $\mathcal L$:
\begin{align}\label{eqn:loss}
\mathcal L &= \log p(X) = \log \prod_n p(\vx_n) = \log \prod_n \sum_k p(\vx_n, z_n{=}k)\nonumber\\
&= \log \prod_n \sum_k \pi_k \mathcal N_k(\vx_n)= \sum_n \log \sum_k \pi_k \mathcal N_k(\vx_n).
\end{align}
The function $\mathcal{L}$ contains only observable quantities and is a suitable loss function for optimization.
At stationary points of \cref{eqn:loss}, the component weights represent the average responsibilities: $\pi_k = \mathbb E_n \gamma_k(\vx_n)$. 
For sampling, one therefore draws a latent variable sample $\hat z\sim\mathcal M(\vpi)$ from a multinomial distribution parameterized by the $\vpi$, 
and then draws a sample $\hat \vt \sim \mathcal N_{\hat z}$ from the component density $\mathcal N_{\hat z}$.
\subsection{DCGMM basics}
The basic data format in a DCGMM instance (see \cref{fig:architecture}) are four-dimensional NHWC tensors. For conciseness, we will
omit the batch dimension (N) from all formulas. 
We will denote the dimensions of the current layer $L$ as $H{,}W{,}C$ and those of the preceding layer $L{-}1$ as $H'{,}W'{,}C'$. Lower-case indices are used similarly, and are sometimes grouped into tuples $\vec m{=}[h{,}w{,}c]^T$ or $\vec m'{=}[h'{,}w'{,}c']^T$ for brevity.

Contrary to DNNs, DCGMMs have two operational modes, see \cref{fig:architecture}: \textit{forward} for density estimation, and \textit{backwards} for sampling.
In forward mode, each layer with index $L{\ge}1$ receives input from the preceding layer $L{-}1$, and generates \textit{activities}  $\mA^{(L)}{\in}\mathbb R^\text{H,W,C}$ which serve as inputs to the subsequent layer $L{+}1$.
As per the usual convention, $L{=}0$ denotes the data samples themselves.
In backwards mode, each layer $L$ receives a \textit{control signal} $\mT^{(L)}{\in}\mathbb R^\text{H,W,C}$ from layer $L{+}1$ and produces another control signal $\mT^{(L{-}1)}{\in}\mathbb R^\text{H',W',C'}$ for layer $L{-}1$.
We define four layer types, see also \cref{app:dcgmm:layers}: \textit{folding} layers $F(f{,}\Delta)$ implementing half-convolutions, CNN-like \textit{max-pooling layers} $P(f{,}\Delta)$ that include a backwards mode, \textit{convolutional GMM layers} $G(K)$ and linear classification layers $C(S)$. 

In the following text, we will discuss how the different  DCGMM layer types implement the computation of loss functions (where applicable), activities and control signals. 
\subsection{Convolutional Gaussian Mixture Layer}\label{sec:cgmm}
cGMM layers are realized by GMMs, slightly modified to a convolutional formulation.
As with any GMM (see \cref{sec:intro:gmm}), this layer type is defined by $K$ weights $\pi_k^{(L)}$, centroids $\vmu_k^{(L)}$ and covariance matrices $\mP_k^{(L)}$.
\smallskip\par\noindent\textbf{Forward mode} Activities $\mA^{(L)}$ of the layer are computed as a function of preceding layer activities $\mA^{(L{-}1)}$. Specifically, activities are realized by GMM posteriors, see \cref{sec:intro:gmm}. 
Since the layer is convolutional, the posteriors are computed from the channel content $A^{(L-1)}_{hw,:}$ at every position $h$,\,$w$ in the tensor of input activities $\mA^{(L{-}1)}$, see \cref{fig:convGMM}:
\begin{align}
P^{(L)}_{hw,k} \big(\mA^{(L-1)}\big) = p\big(A^{(L-1)}_{hw,:}\big)\\
A_{hw,k}^{(L)} \big(\mP^{(L)}\big) = \gamma_k\big(P^{(L)}_{hw,:}\big)
\end{align}
\smallskip\par\noindent\textbf{Backwards mode} In this mode, the cGMM layer generates a control signal $\mT^{(L-1)}$ by sampling at every position $h$,$w$, see \cref{sec:intro:gmm}. Instead of the parameters $\vec \pi$, sampling is governed by an hierarchical prior: the up-stream control signal $\mT^{(L)}$. In analogy to \cref{sec:intro:gmm}, we can write: $T^{(L{-}1)}_{hw,:} \sim \mathcal N_{\hat Z^{(L{-}1)}_{hw}}$, with $\hat Z^{(L{-}1)}_{hw} \sim \mathcal M\big(T^{(L)}_{hw,:}\big)$.
In case there is no control signal (layer is topmost DCGMM layer), the $\hat Z^{(L{-}1)}_{hw}$ are drawn from an uniform distribution.
\smallskip\par\noindent\textbf{Loss function} For efficient and numerically stable training, we use an approximation to the GMM log-likelihood (see \cref{sec:intro:gmm}), which we term the \textit{max-component approximation}, see \cite{gepperth2021c}. It is analogous to EM with hard assignments, see, e.g., \cite{Viroli2019}. As stated in \cref{sec:intro_salient}, each cGMM layer has a loss function that is computed and optimized independently of other layers. For a single sample, it reads:
\begin{align}
\mathcal L^{(L)} = \mathbb E_{h,w} \log \text{max}_k P^{(L)}_{hw,k}
\end{align}
An essential point about cGMM layers is that the activities $\mA^{(L)}$ and control signals $\mT^{(L-1)}$ are computed using a {single} set of centroids, weights and covariance matrices at every position. Thus, these quantities are \textbf{shared} in the same way CNN filters are shared across an image. In this way, large images can be described while requiring relatively few trainable parameters. If memory consumption is not an issue, \textbf{independent} parameters can be used as well.

\subsection{Max-Pooling Layer} 
Pooling is governed by a \textit{kernel size} $f$ and a \textit{stride} $\Delta$, see \cref{fig:convGMM}, where we usually assume non-overlapping pooling: $f {=}\Delta$. We assign to every position $\vec m$ a \textit{receptive field} $\nu(\vec m){=} \{\vec m'{:} h'{\in}[h\Delta{,} h\Delta{+}f[, w'{\in}[w\Delta{,}w\Delta{+}f[, c'{=}c \}$.
In forward mode, this layer type behaves exactly like a CNN max-pooling layer, see \cref{fig:convGMM}: 
$A^{(L)}_{\vec m} = \text{max}_{\vec m'\in\nu(\vec m)} A_{\vec m'}^{(L-1)}$.
In backwards mode, the pooling layer aims at reconstructing a tensor that would have resulted in the provided control signal. We choose to perform \textit{upsampling} here: $T^{(L-1)}_{\vec m'\in\nu(\vec m)}{=}T^{(L)}_{\vec m}$. 
The non-uniqueness of the this mapping must be counteracted by sharpening, see \cref{sec:sampling}.

\subsection{Folding Layer}
\par\noindent\textbf{Forward mode} Folding layers perform a half-convolution on preceding layer activities governed by a \textit{kernel size} $f$ and a \textit{stride} $\Delta$, see also \cref{fig:convGMM}. 
No computation is performed, just a remapping of activities, from position $\vec m'(\vec m)$ in layer $L{-}1$ to $\vec m$ in layer $L$, see \cref{app:m} for details on this relation: $A^{(L)}_{\vec m}{=} A^{(L-1)}_{\vec m'(\vec m)}$.
\par\noindent\textbf{Backwards mode} Inverting the one-to-many mapping $\vec m'(\vec m)$ can be done in $J$ different ways $\vec m_j(\vec m')$. To obtain a control signal, 
we average over all possibilities: $T^{(L{-}1)}_{\vec m'}{=}\frac{1}{J}\sum_j^J T^{(L)}_{\vec m_j(\vec m')}$. 
\subsection{Classification Layer}\label {layers:classification}
This layer implements linear classification for $S$ classes.
Trainable parameters are the weight matrix $\mW$ and the bias vector $\vb$. 
In forward mode, it generates per-class probabilities as $P^{(L)}_s(\vx) = S_s\left(\text{flatten}\big(A^{(L-1)}\big) W{+}\vb\right)$, where $S_s$ represents component $s$ of the softmax function. In backwards mode, it performs an approximate inversion of this operation: $\mT^{(L{-}1)} = \text{reshape}_{H'W'C'}W^T \big(\log(\mT^{(L)}) - \vb + \text{c}\big)$.
The control signal $\mT^{(L)}$ must contain a one-hot encoding of the class that should be generated. The constant $c$ that arises due to the ambiguity in inversing the softmax must be chosen such that the control signal is positive, as it must be if it is to represent GMM posteriors. Classification layers optimize an independent cross-entropy loss function.
%
%
\subsection{End-to-end DCGMM training}\label{sec:end2end}
A defining characteristic of DCGMMs is the fact that each trainable layer 
optimizes its own loss function. We propose a training scheme where all layers are optimized in parallel. To avoid convergence problems, we start adapting the trainable layer $L$ at time $\delta^{(L)}$, which increases 
with the position in the hierarchy. Thus, lower layers achieve some convergence before higher layers start their adaptation, which works well in practice, for small delays $\delta^{(L)}$.

Training cGMM layers is performed by SGD from random initial conditions as detailed in \cite{gepperth2021c}.
As explained in \cite{gepperth2021c}, SGD parameters are very robust and are kept constant throughout all experiments in this article.
\subsection{Density estimation and outlier detection}\label{sec:methods:od}
These are the principal functions of the forward mode, see \cref{fig:architecture}. In contrast to, e.g., deep MFA or SPN instances, see \cref{sec:relwork}, where sample probability is expressed by the root node, any cGMM layer $L$ expresses sample probability by its log-likelihood $\mathcal L ^{(L)}$.
Lower layers usually model small image patches, whereas higher ones capture the global image structure. We evaluate in \cref{sec:exp:outliers} which cGMM layers are best suited for outlier detection.
%
\begin{figure}[t]
\centering
\includegraphics*[page=2,viewport=0in 7.05in 6.in 8.4in, width=0.5\textwidth]{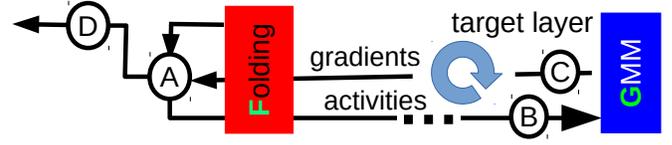}
\caption{\label{fig:sharpening}
The gradient-based sharpening procedure, see text for details. \circled{A} Control signal sampled from folding layer \circled{B} forward propagation of control signal to chosen target GMM layer \circled{C} back-propagated gradients applied for modifying control signal \circled{D} final control signal passed to layer $L{-}1$.
}
\end{figure}
\subsection{Sampling and sharpening}\label{sec:sampling}
Sampling is performed in backwards mode, see \cref{fig:architecture}. Triggered by a top-level control signal, 
each layer $L$ operates in backwards mode, generating control signals $\mT^{(L{-}1)}$ until the sampling result is read out for $L{=}0$. Just as density estimation, sampling is efficiently conducted in mini-batches. As detailed in \cite{gepperth2021a}, control signals can be thresholded and renormalized to contain only the $S$ highest values, which controls the variability of sampling.

Max-pooling and folding layers perform forward transformations that are not invertible. In backwards mode, this means that they can generate control signals that differ from the true data statistics. 
To rectify this, we first observe that up-stream cGMM layers \enquote {know} the statistics of their inputs through training.
Additionally, folding layer outputs contain redundancies since their receptive fields usually overlap, which can be exploited.

We therefore propose a statistics-correcting \textit{sharpening} procedure applied at each folding layer $L$, see \cref{fig:sharpening}. 
First, we operate the folding layer in backwards mode to obtain the initial control signal $\mT^{(L-1)}(i{=}0)$.
We then select a \textit{target layer} $L'{>}L$ (usually the next-highest cGMM layer) and 
perform gradient ascent on $\mT^{(L-1)}(i)$ for $I$ iterations with step size $\epsilon_\text{sh}$. Goal is to maximize the target layer loss $\mathcal L^{(L')}(i)$ obtained by forward-propagating $\mT^{(L{-}1)}(i)$. In this fashion, we obtain a statistics-corrected control signal $\mT^{(L{-}1)}{\equiv}\mT^{(L{-}1)}(i{=}I)$.
%
%
\subsection{Tractable Inference: in-painting}\label{sec:variants}
In in-painting, we present a partially occluded image and ask the DCGMM instance to complete the missing part.
To achieve this, we perform a forward pass up to the topmost cGMM layer $X$, and then a backwards pass with control signal $\mT^{(X)}{=}\tA^{(X)}$. 
The sampling result for the unknown image part is used to complete the input sample.

\section{Methods: models and parameters}\label{sec:sota}
\smallskip\par\noindent\textbf{DCGMM}
We define 7 DCGMM instances which are summarized in \cref{app:dcgmm}. They differ in depth and by their use of stepped folding --vs-- max-pooling.
The principal hyper-parameter is the number of components in cGMM layers, which are chosen such that a batch size of 100 remains feasible. 
In some experiments, we disable parameter sharing between input positions for specific layers.
Learning rates and other SGD-related parameters are kept as described in \cite{gepperth2021a,gepperth2021c}. Since these parameters do not appear to be task-dependent, 
we do not vary them in the presented experiments. The training delay (see \cref{sec:end2end}) for cGMM layer $L$ is set to $\delta^{(L)}{=}0.1L$, where $L{\in}\{1,2,\dots\}$ indicates the number of cGMM layers below the current one.
\smallskip\par\noindent\textbf{RAT-SPN}
We implemented RAT-SPNs as described in the original publication \cite{peharz2020random} using the implementation proposed in \cite{peharz2020random}. As in \cite{peharz2020einsum}, 
for every experiment we vary the following parameters: number of distributions per leaf region/number of sums per sum node $I,S\in\{5, 20,40\}$,  split depth $D\in\{1,5,9\}$ and number of repetitions
$R\in\{10,25,40\}$. To speed up training, we employ binomial leaf distributions. Training is conducted for 25 epochs. Variances are constrained as in \cite{peharz2020random}.
\smallskip\par\noindent\textbf{Deep generalized convolutional SPNs (DGCSPNs)}
We implemented the convolutional architecture for generative experiments from the original publication \cite{wolfshaar2020deep} using \textit{libspn-keras}. The number of sums per sum layer is varied as $S\in\{16,32,64\}$. Training is conducted for 15 epochs. 
Accumulators are initialized by a Dirichlet distribution with $\alpha=0.1$, the number of normal leaf distributions is set to $4$, and centroids are initialized from training data as described in \cite{wolfshaar2020deep}, 
where it is also suggested that variances be kept constant at 1.0 for all normal leaves.
\smallskip\par\noindent\textbf{Poon-Domingos SPN architecture}
As an SPN baseline, we use PD-SPN, a very simple instance of the Poon-Domingos architecture as described in \cite{peharz2020random} using the implementation proposed in \cite{peharz2020einsum}. As in \cite{peharz2020einsum},
we use $\Delta=\{8\}$ for SVHN and $\Delta=\{7\}$ for MNIST, along the horizontal dimension only. 
We vary the number of distributions per leaf region/number of sums per sum node: $I,S\in\{5, 20,40\}$. 
\section{Experiments}
Compared models are DCGMM, RAT-SPN, PD-SPN and DGCSPN, using parameter (ranges) as given in \cref{sec:sota}. 
All experiments were performed on a cluster of 50 off-the-shelf PCs using nVidia GeForce RTX 2060 GPUs, and our own TensorFlow-based implementation of DCGMMs (see \cref{app:libraries} for details).
Generally, we repeat each experiment 5 times with different initializations. When grid-searching for feasible parameters, we use the averaged
performance measures to identify the best settings. Where feasible, we report mean and standard deviations for experiments.

\subsection{DCGMM training dynamics}
%
\begin{figure}
\centering
\includegraphics[width=0.49\linewidth]{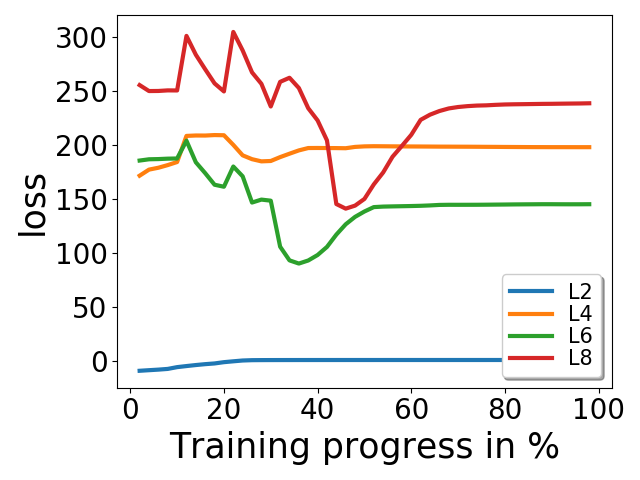}
\includegraphics[width=0.49\linewidth]{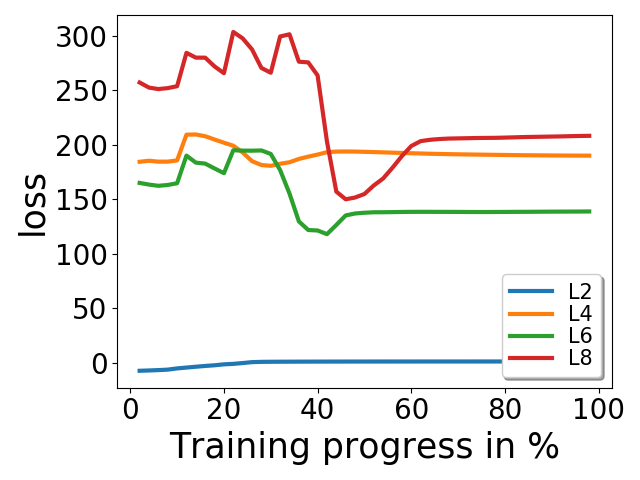}
\caption{\label{fig:exp:dyn} Test losses for all 4 cGMM layers in the deep DCGMM-F instance for MNIST(left) and FashionMNIST(right). Please note that individual cGMM layers' losses have different ranges, and that they are maximized (not minimized) by DCGMM training! 
}
\end{figure}
%
We train the DCGMM-F instance, see \cref{app:dcgmm},
on the full MNIST and FashionMNIST datasets and record the test losses for each  cGMM layer. 
cGMM Parameters are clamped for a certain percentage of training time in different cGMM layers, see \cref{sec:sota}. 
As we can observe in \cref{fig:exp:dyn}, cGMM layer converge sequentially in the order of the hierarchy.
Initially, losses in layers L4, L6 and L8 decrease due to parameter adaptation in lower layers, before increasing once the latter have become stationary. The fact that final losses are sometimes lower than initial ones underscores that they are not global measures for the whole DCGMM.
The individual losses, as depicted in \cref{fig:exp:dyn}, always converge to the values obtained by a layer-by-layer training scheme.
\subsection{DCGMM outlier detection experiments}\label{sec:exp:outliers}
%
\begin{table}
\centering
\begin{tabular}{|l|cc|l|cc}

 ID & \multicolumn{1}{c}{\scriptsize{MNIST}} &  \multicolumn{1}{c|}{\scriptsize{FashionMNIST}} & ID & \multicolumn{1}{|c}{\scriptsize{MNIST}} &  \multicolumn{1}{c}{\scriptsize{FashionMNIST}} \\
  \hline
A & \textbf{95.4}${\pm}$0.2         & 62.2${\pm}$0.3       & E & 92.8${\pm}$0.4  & \textbf{74.9}${\pm}$0.2 \\
B & 93.3${\pm}$0.5                  & 72.2${\pm}$0.2           & F & 84.1${\pm}$0.4  & 68.1${\pm}$0.1 \\
C & 94.2${\pm}$0.2                  &        {74.3}${\pm}$0.1  & G & 77.4${\pm}$0.3  & 62.7${\pm}$0.1 \\
D & 92.5${\pm}$0.3                  & 68.1${\pm}$0.1          & - & - &  -
\end{tabular}
\caption{\label{tab:od} Outlier detection by the top layers of various DCGMM instances, see \cref{app:dcgmm}, measured by the mean AUC (in \%). }
\end{table}
The goal of this experiment is to assess the outlier detection capabilities of DCGMMs. In particular, we investigate which DCGMM architectures are most suited for this purpose, and 
which cGMM layer the outlier detection should be based on. 
We construct outlier detection problems from classes 1--9 (inliers) --vs-- 0 (outliers) of the MNIST and FashionMNIST datasets. 
Test losses on inlier and outlier classes are recorded for all layers in a DCGMM instance after training on the inliers. 
Outlier detection is evaluated separately based on each cGMM layer's loss.
By varying the threshold for outlier detection (see \cref{sec:methods:od}), we obtain ROC-like outlier detection plots, 
see \cref{app:outliers}. The area-under-the-curve (AUC) is used as a quality measure.
Results are summarized in \cref{tab:od}. We find that the highest cGMM layers are most suited for outlier detection, so we report performance only for these layers. For FashionMNIST, the depth of a DCGMM instance increases its outlier detection capacity. For MNIST, the \enquote{flat} DCGMM-A performs best, but MNIST might really be to simple a benchmark, and the deep instances have similar performance. Results with different outlier classes are comparable.
\subsection{Outlier detection: model comparison}\label{sec:exp:comp}
\begin{table}
\centering 
\begin{tabular}{c|cc}
ID        &            {\scriptsize{MNIST AUC in \%}} & {\scriptsize{FMNIST AUC in \%}}\\
\hline
DCGMM-E   & \textbf{92.8}$\pm$ 0.4   & \textbf{74.9} $\pm$ 0.2\\
RAT-SPN   & 91.8 $\pm$ 0.7  & 39.2 $\pm$ 0.4 \\ 
DGCSPN   &  90.6 $\pm$ 0.7 & 57.1 $\pm$  0.9  \\    
PD-SPN    &  91.2$\pm$ 0.6  &   48.5 $\pm$ 1.7   \\ 
\hline
\end{tabular}
\caption{\label{tab:od} Outlier detection for DCGMM-E and SPN models for MNIST and FashionMNIST, quantified by the AUC measure.
}
\end{table}
Here, we compare the outlier detection capacity of DCGMM-E, the best DCGMM instance from \cref{sec:exp:outliers}, to various RAT-SPN, PD-SPN and DGCSPN instances, using the same procedures.
To find the best parameters for each SPN type, a grid search is conducted over parameters ranges as stated in \cref{sec:sota}, repeating each experiment 5 times with identical parameters. 
Best outlier detection capacities, measured as in \cref{sec:exp:outliers}, are reported in \cref{tab:od}. We observe a clear edge for DCGMM-E, in particular for FashionMNIST.

\subsection{DCGMM sampling and sharpening}\label{sec:exp:sampling}

\begin{figure*}
\centering
\includegraphics[width=0.24\textwidth]{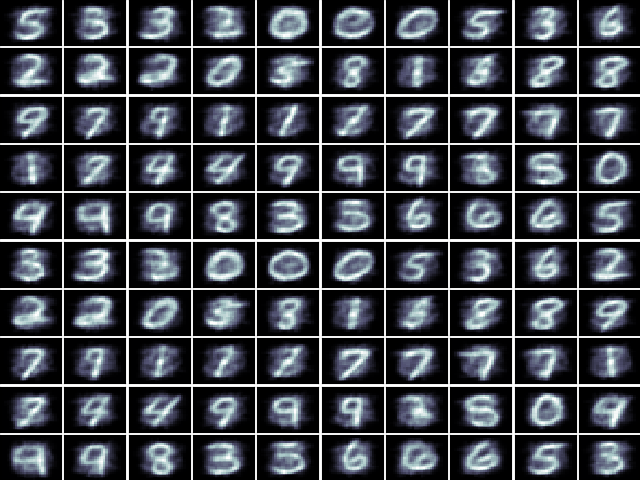}
\includegraphics[width=0.24\textwidth]{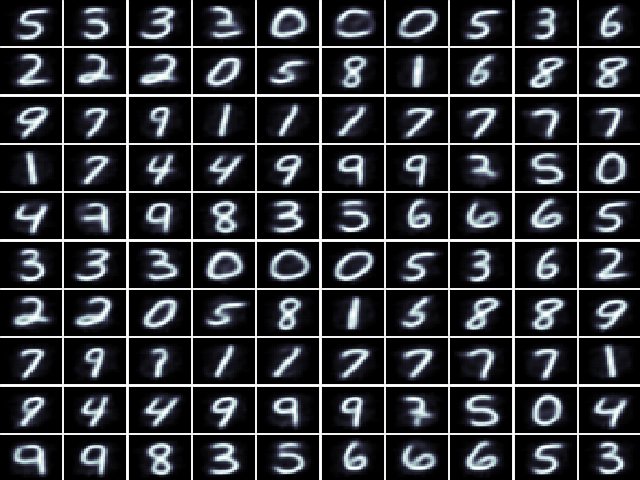}
\includegraphics[width=0.24\textwidth]{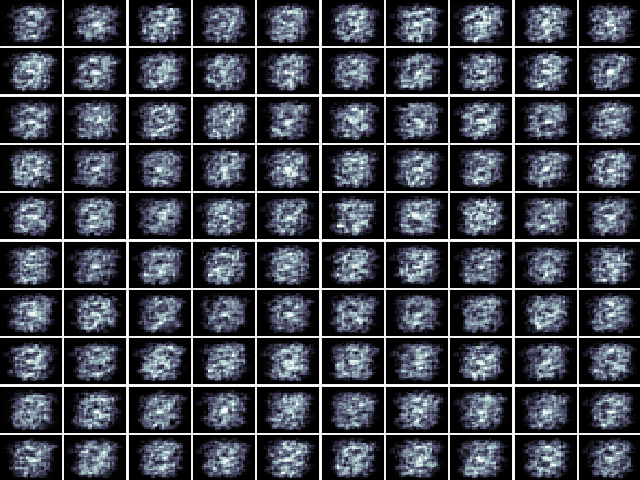}
\includegraphics[width=0.24\textwidth]{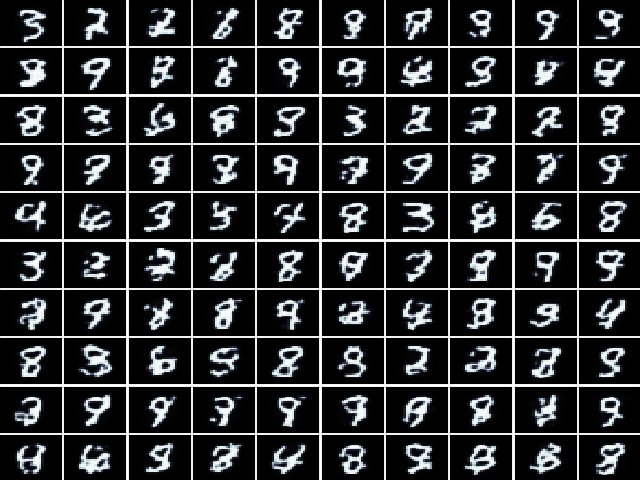}
\caption{\label{fig:exp:sharp} Samples from DCGMMs with pooling. Left to right: DCGMM-C (no sharpening), DCGMM-C (sharpening), DCGMM-D (no sharpening), DCGMM-D (sharpening). The most beneficial effect of sharpening is observed for the shallow DCGMM-C instance.
}
\end{figure*}
To demonstrate how the sharpening procedure of \cref{sec:sampling} improves sampling through non-invertible operations, we train 
DCGMM instances with max-pooling (C and D, see \cref{app:dcgmm}), to sample from MNIST and FashionMNIST. Then, we compare sampling results with and without sharpening, see \cref{fig:exp:sharp}.
When sharpening is used, target layers are always the next-highest cGMM layers, and gradient ascent is performed for $I=300$ iterations using a step size $\epsilon_\text{sh}{=}1.0$.
\Cref{fig:exp:sharp} shows that strong blurring effects occur without sharpening, due to ambiguities in inverting max-pooling layers. In contrast, sharpening removes these ambiguities while maintaining diverse samples.
This works best for DCGMM-B (one max-pooling layer), whereas more max-pooling steps seem to destroy too much information, leading to frayed-looking samples. 
We conclude that \enquote{softer} strategies than max-pooling may be required for sampling with really deep DCGMMs.
The corresponding figures for FashionMNIST are given in \cref{app:sharpening} and corroborate this view.

\subsection{Sampling: visual model comparison}\label{sec:sampling:comp}
%
\begin{figure*}
\centering
\includegraphics[width=0.24\textwidth]{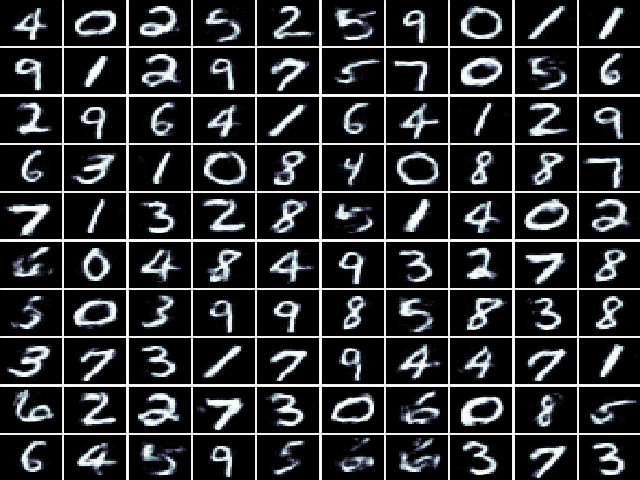}
\includegraphics[width=0.24\textwidth]{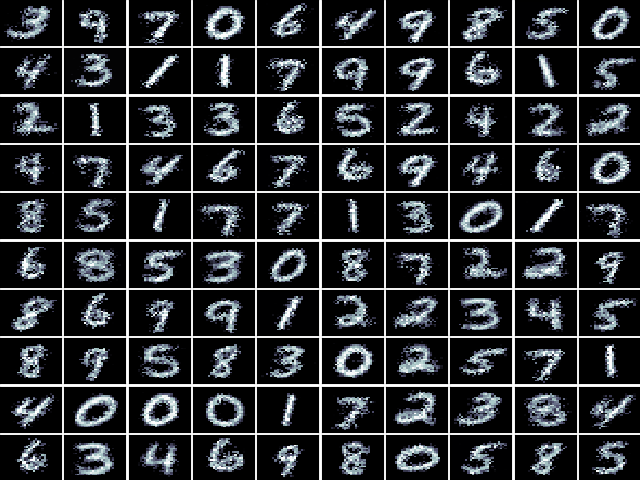}
\includegraphics[width=0.24\textwidth]{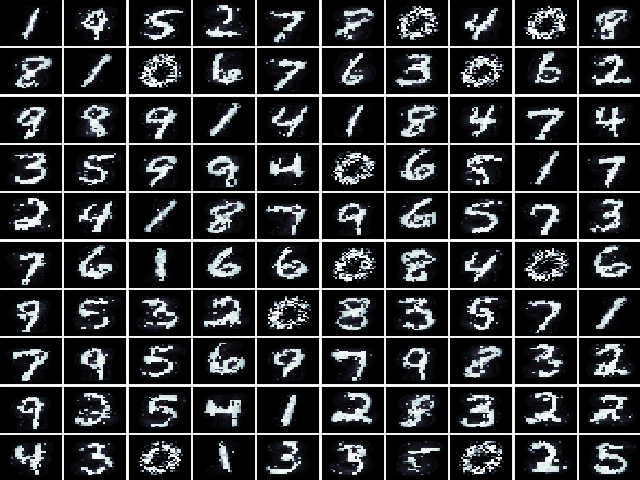}
\includegraphics[width=0.24\textwidth]{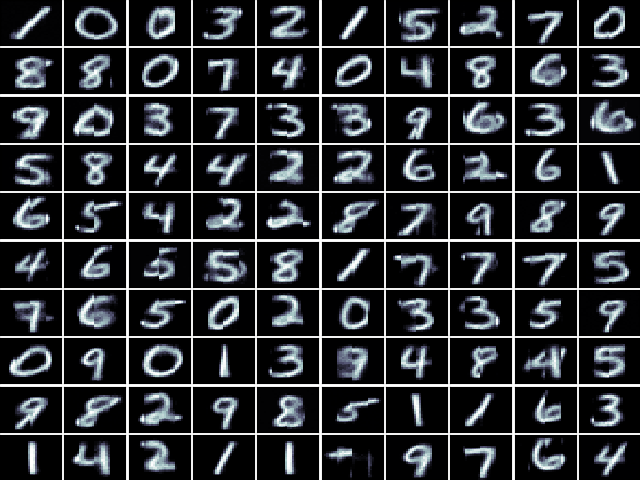}
\caption{\label{fig:samples:vis}
Visual comparison of sampling results. From left to right: DCGMM-F, RAT-SPN, PD-SPN, DGCSPN. 
}
\end{figure*}
In this experiment, we perform a visual comparison of samples generated from DCGMM and the SPN models described in \cref{sec:sota}.
SPN models are used in the configurations given in the literature (see \cref{sec:sota}) and the best parameters found by grid-search in \cref{sec:exp:comp}. Since the configurations in the literature are tailored to certain datasets, we restrict this investigation to the MNIST and FashionMNIST datasets.
For DCGMM, we selected DCGMM-F, see \cref{app:dcgmm}, an instance without max-pooling, since this generally leads to 
more realistic samples, see \cref{sec:exp:sampling}. 

Typical samples generated by these instances are shown in \cref{fig:samples:vis}. We observe that the DCGMM samples generally look smoother and more realistic than SPN-generated ones. Instances from other DCGMM instances, and the corresponding FashionMNIST samples, are shown in \cref{app:samples-fmnist}.
\subsection{Sampling: quantitative model comparison}\label{sec:exp:cond}
%
%
\begin{table}
\centering
\begin{tabular}{cc|cc}      
ID    & parameters  & {MNIST acc.} & {FMNIST acc.}\\
\hline
DCGMM-A &   38.416    & 98.2$\pm$ 0.1     & 80.3$\pm$ 2.6    \\
DCGMM-B &   293.657    & 98.7$\pm$ 0.05   & 83.6$\pm$ 1.7 \\
DCGMM-E &    40.850   & 99.1$\pm$ 0.2     & \textbf{94.5} $\pm$ 1.7 \\
DCGMM-F &   50.850    & \textbf{99.9}$\pm$ 0.07     & 89.0 $\pm$ 2.5\\ 
RAT-SPN &   1.187.840   & 99.2$\pm$ 0.1   & 85.4 $\pm$ 1.9    \\
DGCSPN &   7.560.576  & 98.0$\pm$ 0.1     & 80.4$\pm$ 6.0        \\ 
PD-SPN  &    515.683 & 97.2$\pm$ 0.2     &  81.2$\pm$ 2.4         \\ 
\hline
\end{tabular}
\caption{\label{tab:exp:cond}
Sample generation capacity for DCGMM and SPN models as measured by a CNN classifier on generated samples, see text for details. We observe that all models perform very similarly on MNIST, whereas DCGMM-E outperforms SPNs for FashionMNIST. Averages and standard deviations are taken over 5 independent training runs of the CNN classifier. 
}
\end{table}
Since visual appearance can be deceptive, this experiment aims to provide a quantitative comparison of the sampling capacity of DCGMMs and SPNs, relying on 
MNIST and FashionMNIST. We first train a CNN classifier on both datasets to deliver state-of-the-art accuracy for CNNs (details in \cref{app:cnn}).
DCGMMs and SPNs are separately trained on each class in both datasets, and asked to generate 1000 samples from all 10 classes. The quality measure is the CNN's classification accuracy on the generated samples. 
Results are shown in \cref{tab:exp:cond}. We observe that DCGMM sampling seems to produce samples that more accurately match the real data than SPN-based models.
\subsection{Sampling for complex visual problems}\label{sec:exp:svhn}
%
\begin{figure*}
\centering
\includegraphics[width=0.24\textwidth]{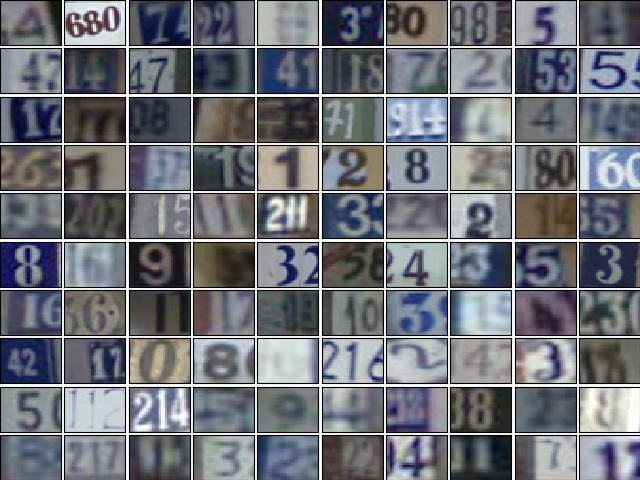}
\includegraphics[width=0.24\textwidth]{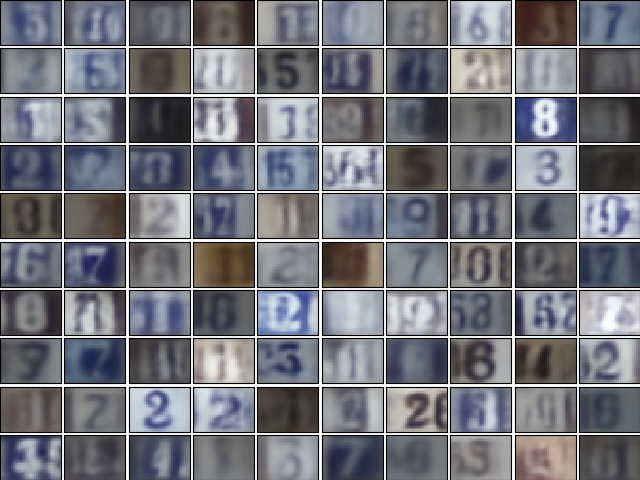}
\includegraphics[width=0.24\textwidth]{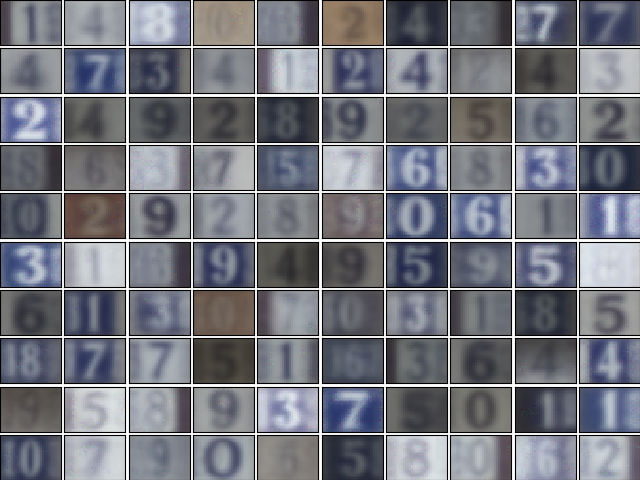}
\includegraphics[width=0.24\textwidth]{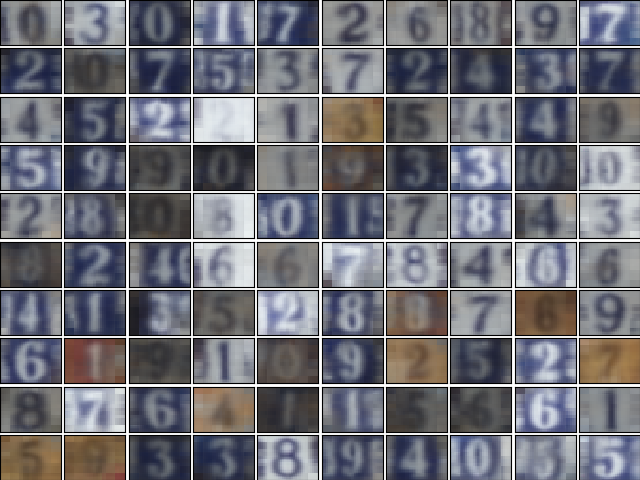}
\caption{\label{fig:sampling:svhn} SVHN sampling. Left to right: a) SVHN samples b) VAE c) DCGMM-A d) DCGMM-B.
}
\end{figure*}
Here, we demonstrate that DCGMMs can be trained on complex and high-dimensional visual problems like SVHN, and generate high-quality samples when compared to SPNs. 
DCGMM instances A and B (the latter with the higher cGMM layer in \enquote{independent} mode, see \cref{sec:cgmm}) are separately trained on each SVHN class, and then used for sampling. 
We observe that the deeper DCGMM-B instance generates more diverse images, and that DCGMM-B samples even compare favorably to VAE samples (details in \cref{app:vae}).
\subsection{Inference: in-painting} 
\begin{figure}
\centering
\includegraphics[width=0.49\linewidth]{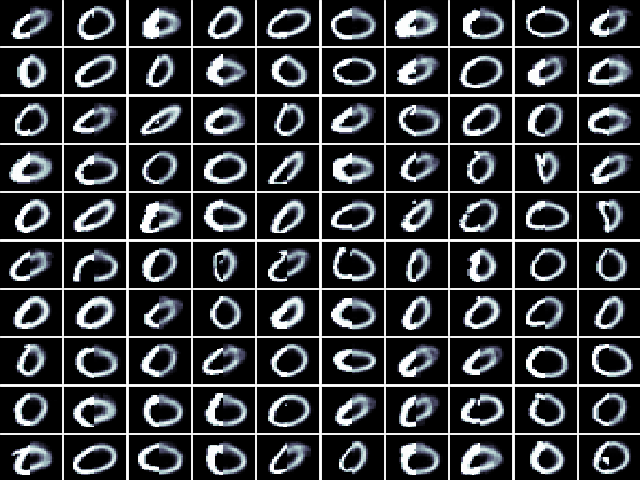}
\includegraphics[width=0.49\linewidth]{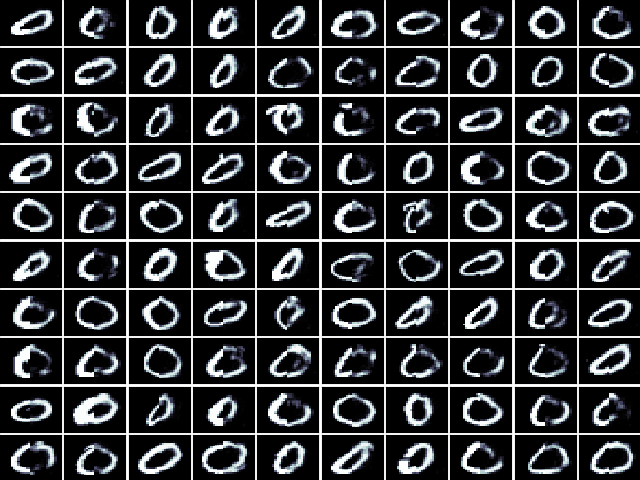}
\caption{\label{fig:inpainting} Examples of in-painting the erased right half of an image. Left: DCGMM-A, right: DCGMM-F.
}
\end{figure}
We perform in-painting (see \cref{sec:variants}) on MNIST samples from which the right half was erased. 
In-painting has the same complexity as sampling, i.e., linear in the number of cGMM layers. Results for the shallow  DCGMM-A and the deep DCGMM-F instance are shown in \cref{fig:inpainting}. We observe that completion fits the original image better with a deep DCGMM instance. This is natural since DCGMM-A is essential a vanilla GMM and can just replicate its limited set of components, with
added Gaussian noise. A deep DCGMM instance, in contrast, exhibits much greater variability since each cGMM layer adds (guided) randomness in the choice of 
the component to sample from.
\subsection{Generative-discriminative learning}\label{sec:exp:class}
\begin{table}
\centering 
\begin{tabular}{c|cc}
model        &            {\scriptsize{MNIST acc. in \%}} & {\scriptsize{FMNIST acc. in \%}}\\
\hline
DCGMM-B   & 98.0 $\pm$ 0.1   & 89.6 $\pm$ 0.2\\
RAT-SPN   & 98.19  & 89.52 \\ 
DGCSPN   &  98.66 & 90.74  \\    
\hline
\end{tabular}
\caption{\label{tab:class} Classification accuracies in \% obtained for MNIST and FashionMNIST. For RAT-SPNs and DGCSPNs, mean accuracy over 5 runs is taken from \cite{wolfshaar2020deep}.
}
\end{table}

This experiment assesses DCGMM classification accuracy on MNIST and FashionMNIST by adding a top-level classification layer
to the shallow DCGMM-B instance (see \cref{tab:dcgmms}). To boost classification performance, we use activities of the $2$ highest cGMM layers as input to the classifier layer. 
We find that DCGMM-B classification accuracy is similar but slightly inferior to DGCSPN and RAT-SPN from the literature, see \cref{tab:class}, and that deeper DCGMM instances consistently gave worse results.
\section{Principal Conclusions from Experiments}
\smallskip\par\noindent\textbf{Sample probability} is best expressed by the topmost cGMM layer loss, see \cref{sec:exp:outliers}. 
This is an important result, since DCGMMs optimize several loss functions, each of which expresses a different local model of the data.

\smallskip\par\noindent\textbf{Max-pooling} produces excellent results for outlier detection, see \cref{sec:exp:outliers},
and is feasible for sampling in shallower DCGMMs due to sharpening, see \cref{sec:exp:sampling}). However, 
in deep DCGMM instances, the information loss seems too severe for good sampling performance. Here, stepped, overlapping half-convolutions are found to be a better choice.

\smallskip\par\noindent\textbf{Parameter sharing} is very effective for reducing the number of model parameters especially in lower layers, and does not seem to impair sampling performance (see \cref{sec:exp:svhn}). 
This is probably because the local \enquote{visual alphabet} (see \cref{app:alphabet}) is nearly position-invariant in lower layers and thus can be shared between positions with little loss.

%
%
\smallskip\par\noindent\textbf{Classification} 
Since DCGMMs optimize independent loss functions, discriminative training of a classification layer does not
influence unsupervised training in lower layers, because no back-propagation is performed. 
Thus, the DCGMM is still be able to generate realistic samples, at the price of similar but slightly inferior classification accuracy, see \cref{sec:exp:class}.
This is in contrast to discriminative SPN training, see \cite{wolfshaar2020deep,peharz2020random}. Here, RAT-SPNs and DGC-SPNs perform classification by optimizing a global cross-entropy loss, but can no longer generate realistic samples.

\smallskip\par\noindent\textbf{Realistic sampling} The DCGMM samples presented in \cref{sec:exp:sampling,sec:exp:svhn} have a clear edge w.r.t. samples produced by SPNs. This shows that linking independent probabilistic descriptions (the cGMM layers) by a deep chain of hierarchical priors is a feasible way to describe complex (image) distributions.
SPN training on complex problems like SVHN is described in \cite{peharz2020einsum}, where the PD-SPN architecture (see \cref{sec:sota}) is trained on preprocessed SVHN data. Due to non-overlapping scopes in the PD architecture, all samples exhibit \enquote{stripe} artifacts which are absent from DCGMM samples, presumably because their scopes \textit{can} overlap.

%
%
%
%
%
\bibliographystyle{abbrv}
\bibliography{ijcnn2021}

\appendix
%
\section{DCGMM architectures}\label{app:dcgmm}
Precise layer configurations of the various DCGMM instances used in this article are given in \cref{tab:dcgmms}. As a rule, cGMM components were chosen as high as possible while respecting memory constraints during training. All cGMM layers are assumed to use parameter sharing, otherwise the number of trainable parameters will be higher. We observe that the number of trainable parameters actually decreases as more cGMM layers are added. As with CNNs, this is because the number of parameters mainly scales with the filter size of folding layers, which can be kept small in deep architectures.
\begin{table}[h!]
\centering
\begin{tabular}{llc}
ID & Configuration & parameters\\
\hline
$A$   & F(28,1)-\textbf (49) & 38416 \\
$B$   & F(8,2)-\textbf G(49)-F(11,1)-\textbf G(49) & 293657 \\
$C$   & F(8,1)-\textbf G(49)-P(2,2)-\textbf G(49)& 243236 \\
$D$ & F(3,1)-\textbf G(25)-P(2,2)-F(4,1)-\textbf G(25)- &      \\
    & P(2,2)-F(5,5)-\textbf G(49)              &  40850\\
$E$ & F(3,1)-\textbf G(25)-F(4,2)-\textbf G(25)-  & \\
    & F(12,1)-\textbf G(49)                       & 186625\\
$F$ & F(3,1)-\textbf G(25)-F(4,2)-\textbf G(25)-F(4,2)-  &   \\
    &\textbf G(25)-F(5,1)-\textbf G(49)        & 50850 \\
$G$ & F(3,1)-\textbf G(25)-P(2,2)-F(3,1)- & \\
    &  \textbf G(25)-P(2,2)-F(3,1)- G(25)-P(2,2)-  & 	 \\
    &  F(2,1)-\textbf G(49) & 16375 \\
\end{tabular}
\caption{\label{tab:dcgmms} Overview of DCGMM configurations used in the experiments. Layer types are F (half-convolution layer), G (cGMM layer) and P(max-pooling layer). 
Optionally, a linear classifier layer can be added at the top of each instance for conditional sampling. 
}
\end{table}
\section{More details on DCGMM layers}\label{app:dcgmm:layers}
\begin{table}
\centering
\begin{tabular}{cc|ccc}
\hline
Layer type & Notation & $H$ & $W$ & $C$\\
\hline
Folding &  $F(f, \Delta)$ & $1{+}\frac{H'-f}{\Delta}$ & $1{+}\frac{W'-f}{\Delta}$ &  $f^2 C'$ \\
Max-Pooling &  $P(f, \Delta)$ & $1{+}\frac{H'-f}{\Delta}$ & $1{+}\frac{W'-f}{\Delta}$ &  $C'$ \\
Classification  & $C(S)$ & 1 & 1 & $S$  \\
cGMM  & $G(K)$ & $H'$ & $W'$ & $K$ 
\end{tabular}
\caption{\label{tab:layers} Overview of DCGMM layer types and their notation. The three rightmost columns indicate the shape of activities in forward mode if the layer receives an input of dimensions $H'{,}W'{,}C'$. Folding and max-pooling layers are parameterized by kernel size $f$ and the stride $\Delta$, GMM layers by the number of components $K$ and classification layers by the number of classes $S$.
}
\end{table}
%
DCGMM layers transform inputs of dimension $H'{,}W'{,}C'$ into activities of dimension $H{,}W{,}C$. Since each layer in forward mode implements a deterministic transformation, the dimensions of activities depend only on the dimensions of the inputs as listed in \cref{tab:layers}.

\section{CNN for measuring conditional sampling}\label{app:cnn}
The CNN classifier used in \cref{sec:exp:cond} has a layer structure as given in \cref{tab:cnn}. It is implemented in TensorFlow2/Keras and trained for 15 Epochs on either MNIST or FashionMNIST, using an Adam optimizer, a learning rate of 0.01 and a batch size of 100.
For MNIST, this is sufficient for state-of-the-art performance ($>$ 99\%), whereas for FashionMNIST, a performance of roughly 91\% can be reached. While this is inferior to the performance obtained by more 
refined models (e.g., \cite{tanveer2021fine}), we accept it here for simplicity, and also because the CNN classifier is just a tool to detect differences in the distributions of real and generated samples.
\begin{table}[h!]
\centering
\begin{tabular}{cccc}
Type & Kernel & Prob. & channels/neurons\\
\hline
Dropout & & 0.1 & \\
Conv/ReLU & 3 & & 64\\
Conv/ReLU & 3 & & 64\\
Pooling & 2 & & \\
Conv/ReLU & 3 & & 64\\
Pooling & 2 & & \\
Conv/ReLU & 3 & & 64\\
Pooling & 2 & & \\
Dense/ReLU & & & 350\\
Dropout & & 0.1 & \\
Dense/ReLU & & & 350 \\
Dense/ReLU &  & & 350\\
Dense/Softmax &  & & 10
\end{tabular}
\caption{\label{tab:cnn} Hyper-Parameters of the CNN used for assessing sampling performance.
}
\end{table}

\section{DCGMM sharpening for FashionMNIST}\label{app:sharpening}
\begin{figure*}[t!]
\centering
\includegraphics[width=0.24\textwidth]{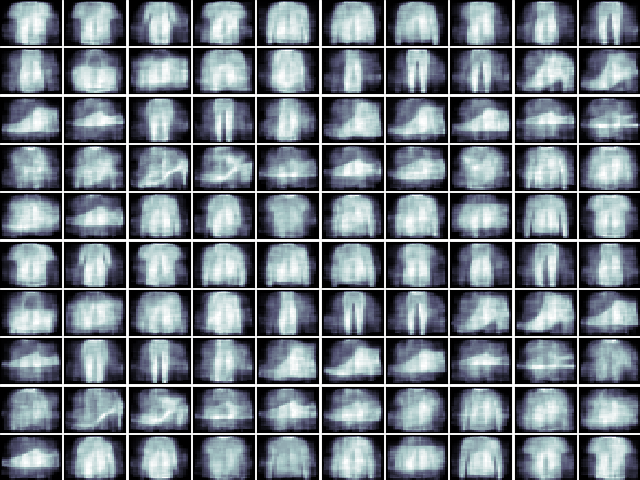}
\includegraphics[width=0.24\textwidth]{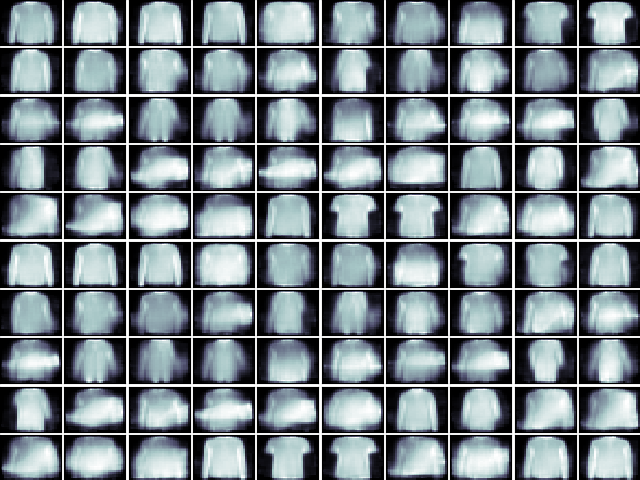}
\includegraphics[width=0.24\textwidth]{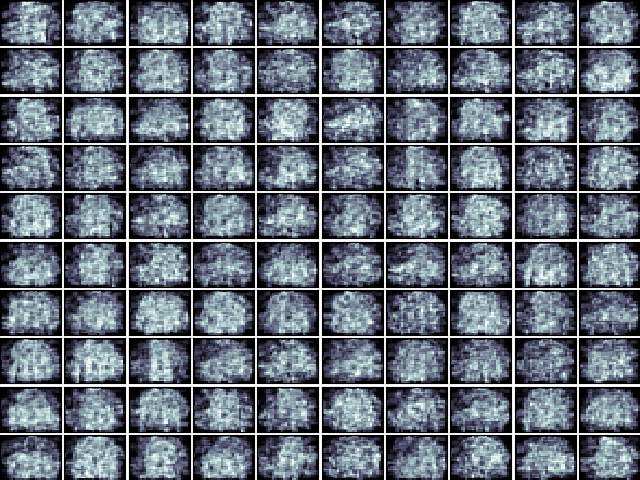}
\includegraphics[width=0.24\textwidth]{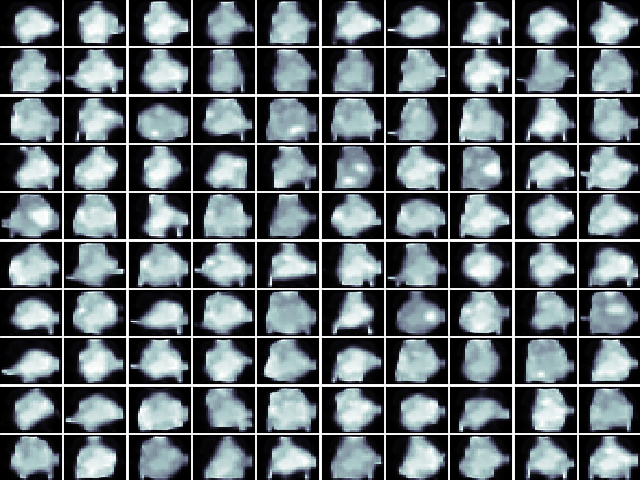}
\caption{\label{fig:exp:fm-sharp} 
Sharpening for DCGMMs with pooling on FashionMNIST. Left to right: DCGMM-C (no sharpening), DCGMM-C (sharpening), DCGMM-D (no sharpening), DCGMM-D (sharpening). The most beneficial effect of sharpening is observed for the shallow DCGMM-C instance.
}
\end{figure*}

Sharpening behaves similarly for the FashionMNIST dataset as it was found for MNIST in \cref{sec:exp:sampling}. Namely, 
the shallower DCGMM-B instance profits strongly from a sharpening through a single max-pooling layer, but we observe deterioration of sampling performance when 
more max-pooling layers are involved, as in instance DCGMM-D. \Cref{fig:exp:fm-sharp} shows this quite nicely.

\section{More details on outlier detection}\label{app:outliers}
\begin{figure*}[h!]
\centering
\includegraphics[width=0.32\textwidth]{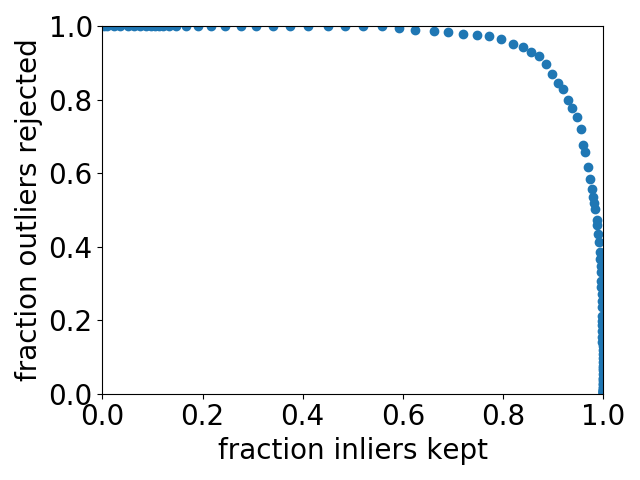}
\includegraphics[width=0.32\textwidth]{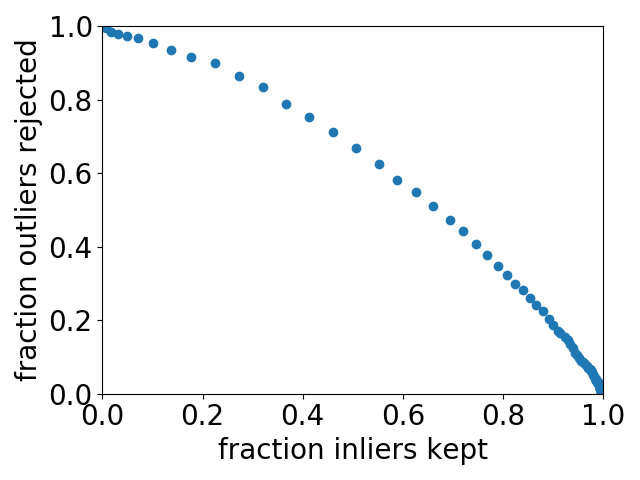}
\includegraphics[width=0.32\textwidth]{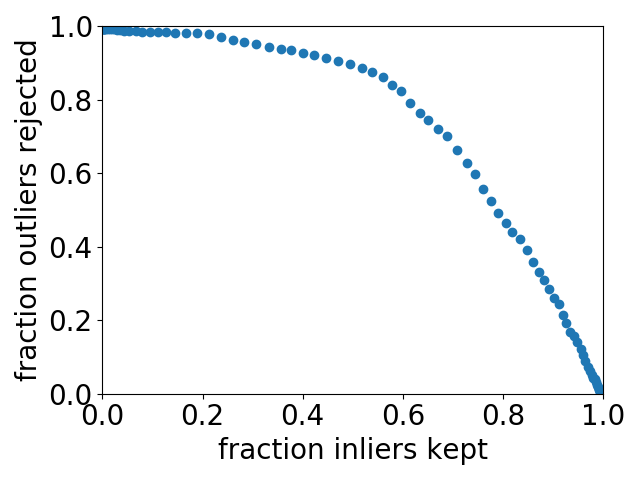}
\caption{\label{fig:app:roc} Examples of ROC-like curves for outlier detection. Left: DCGMM-A (MNIST), middle: DCGMM-A (FashionMNIST), right: DCGMM-E(FashionMNIST). 
The area under these curves is taken to be a measure of outlier detection capacity. 
}
\end{figure*}

Outlier detection is quantified using a ROC-like curve, plotting kept inliers against rejected outliers while varying the separation threshold that
is applied to the log-likelihoods. Typical examples of such curves are shown in \cref{fig:app:roc}. 
\begin{figure*}[t]
\centering
\includegraphics[width=0.31\textwidth]{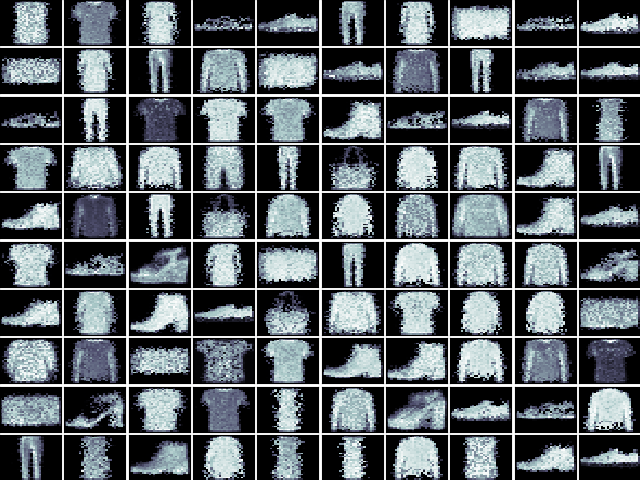}
\includegraphics[width=0.31\textwidth]{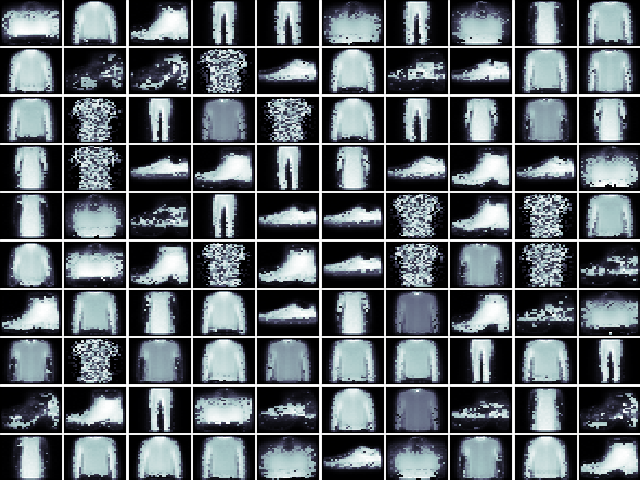}
\includegraphics[width=0.31\textwidth]{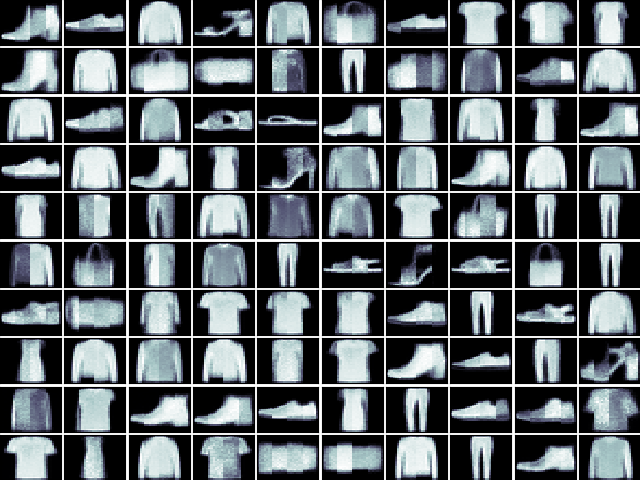}
\caption{\label{fig:samples:extra3} SPN samples for FashionMNIST. From left to right: RAT-SPN, PD-SPN, DGCSPN.
}
\end{figure*}
\section{Variational autoencoder details}\label{app:vae}
\Cref{tab:ae} gives details about the convolutional VAE used to generate SVHN samples. It was trained for 100 epochs on all SVHN classes using the Adam optimizer and a learning rate of $0.0001$. 
\begin{table}[h!]
\centering
\begin{tabular}{cccc}
Type & Kernel & Stride & channels/size\\
\hline
\multicolumn{4}{c}{Encoder} \\
\hline
Conv/ReLU & 3 & 1 & 64\\
Conv/ReLU & 3 & 2 & 128\\
Conv/ReLU & 3 & 2 & 256\\
Conv/ReLU & 2 & 2 & 512\\
Flatten & - & - & 4096 \\
Dense & & & 2*64\\
\hline
\multicolumn{4}{c}{Decoder} \\
\hline
Dense/ReLU & & & 4*4*128 \\
Reshape & & &4,4,128 \\
Conv2D$^T$/ReLU & 4 & 1 & 1024\\
Conv2D$^T$/ReLU & 5 & 1 & 512\\
Conv2D$^T$/ReLU & 4 & 2 & 256\\
Conv2D$^T$/ReLU & 5 & 1 & 128\\
Conv2D$^T$/ReLU & 3 & 1 & 128\\
Conv2D$^T$/ReLU & 3 & 1 & 3\\
\end{tabular}
\caption{\label{tab:ae} Hyper-Parameters of the VAE used for SVHN sampling.
}
\end{table}

\section{Additional sampling results}\label{app:samples-fmnist}
\begin{figure}
\centering
\includegraphics[width=0.23\textwidth]{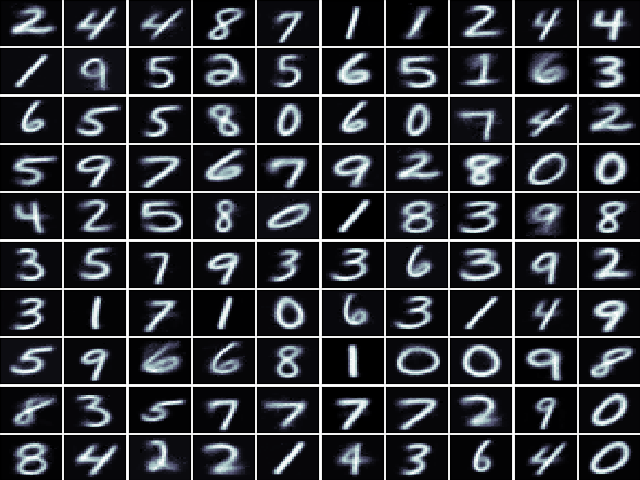}
\includegraphics[width=0.23\textwidth]{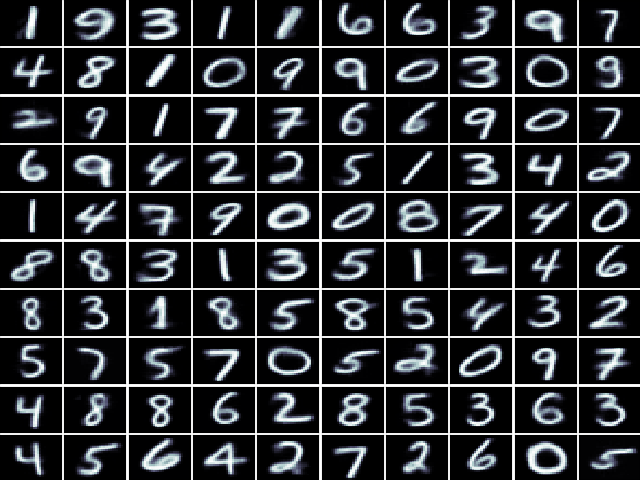}\\
\vspace{0.5em}
\includegraphics*[width=0.23\textwidth]{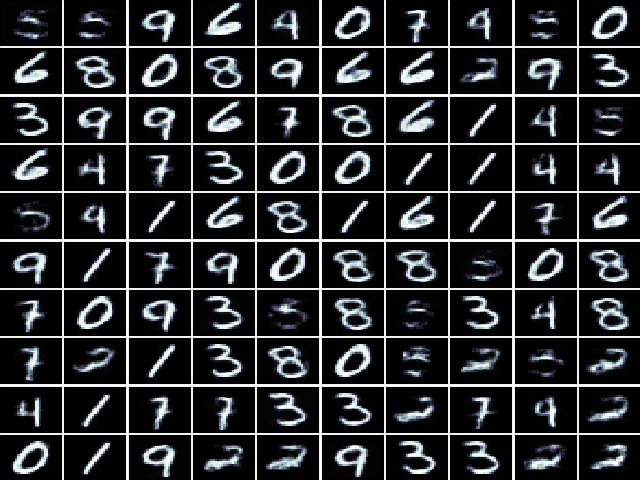}
\includegraphics*[width=0.23\textwidth]{figs/samples-dcgmm-f.png}
\caption{\label{fig:samples:extra1} Samples from several DCGMM instances for MNIST. Upper row: DCGMM-A(left), DCGMM-B(right). Lower row: DCGMM-E(left), DCGMM-F(right).
}
\end{figure}

\begin{figure}
\centering
\includegraphics[width=0.23\textwidth]{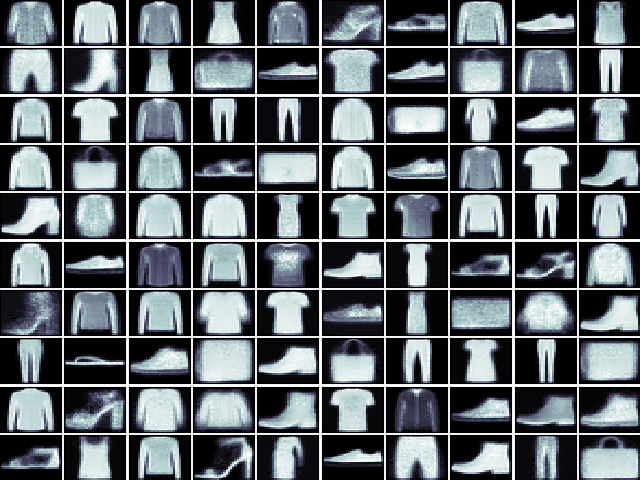}
\includegraphics[width=0.23\textwidth]{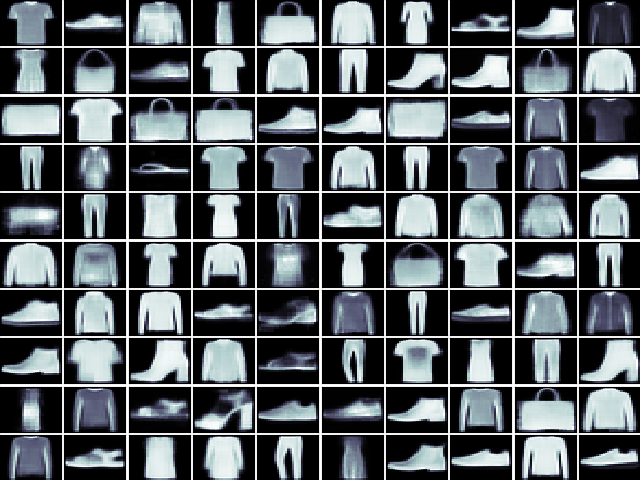}\\
\vspace{0.5em}
\includegraphics[width=0.23\textwidth]{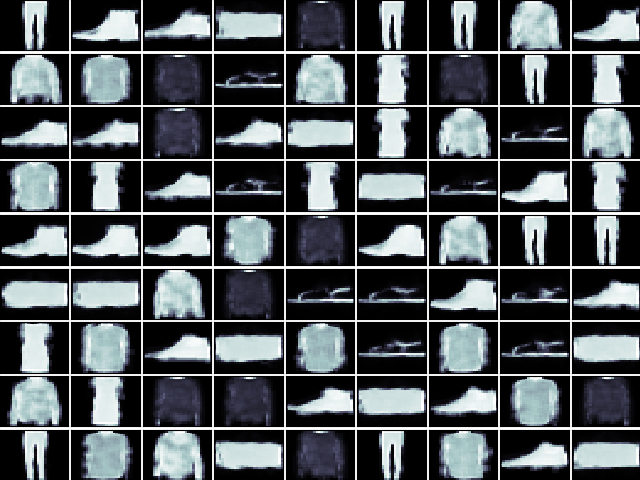}
\includegraphics[width=0.23\textwidth]{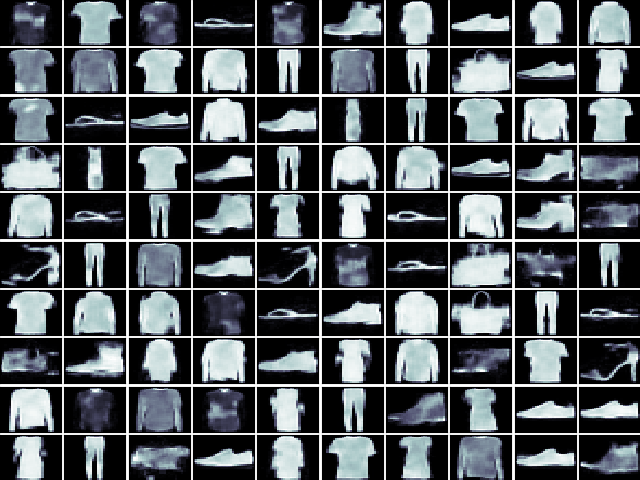}
\caption{\label{fig:samples:extra2} Samples from several DCGMM instances for FashionMNIST. Upper row: DCGMM-A (left), DCGMM-B (right).
Lower row: DCGMM-E (left), DCGMM-F (right).
}
\end{figure}

This appendix gives MNIST sampling results in a more complete fashion, that is, including more DCGMM instances, in \cref{fig:samples:extra1}. 
\Cref{fig:samples:extra2} gives samples from the same DCGMM instances for FashionMNIST, whereas \cref{fig:samples:extra3} shows FashionMNIST samples generated by SPNs.

\begin{figure}
\centering
\includegraphics[width=0.23\textwidth]{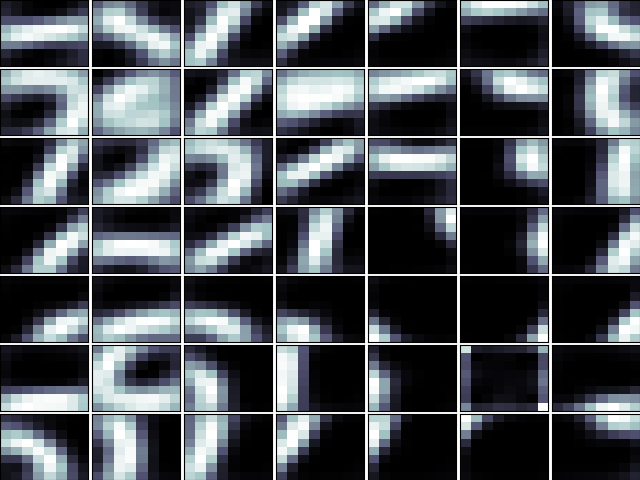}
\includegraphics[width=0.23\textwidth]{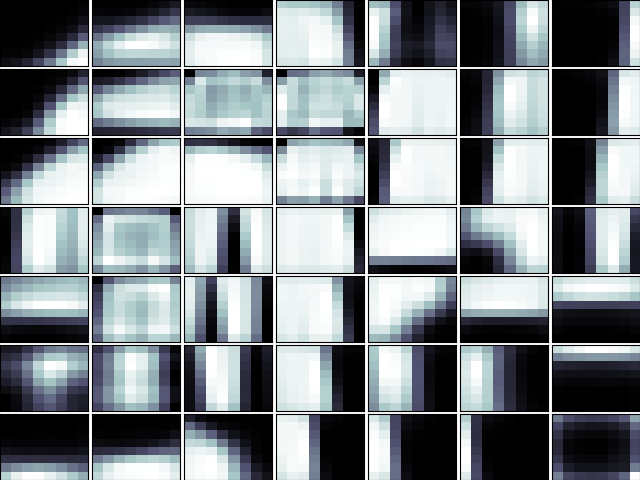}
\caption{\label{fig:alphabet} Visualization of centroids in the lowest cGMM layer of DCGMM-B. Left: training on MNIST, right: training on FashionMNIST.
}
\end{figure}
\section{Visual alphabet}\label{app:alphabet}
The lower cGMM layers of a cGMM instance usually model small image patches extracted by a preceding folding layer. 
With parameter sharing enabled, the cGMM therefore describes all positions within an image using a single set of parameters. 
This makes the most sense for low-hierarchy layers, since local image content tends to be similar across image at small patch sizes.
The cGMM prototypes there form a kind of \enquote{visual alphabet}, a set of centroids that, together, best describe local image content.
We exemplarily show this for MNIST and FashionMNIST by visualizing the centroids of the lowest cGMM layer in instance DCGMM-B, which models
8x8 image patches. We observe in \cref{fig:alphabet} that the basic building blocks of both datasets are faithfully represented.
\section{Libraries and public code}\label{app:libraries}
For implementing RAT-SPN and PD-SPN, we made use of the public code provided under \url{https://github.com/cambridge-mlg/EinsumNetworks} which relies mainly on PyTorch. 
DGCSPNs are implemented using \textit{libspn-keras} which is TensorFlow2-based and can be obtained from \url{https://github.com/pronobis/libspn-keras}. VAEs and CNNs are self-implemented in TensorFlow2/Keras. 
TensorFlow2-Code for DCGMM can be found under \url{https://github.com/anon-scientist/ijcnn22-dcgmm}. Code for selected experiment of this article is available under 
\url{https://github.com/anon-scientist/ijcnn22-experiments}.

\section{Folding layer details}\label{app:m}
The relation between preceding and current layer activities is governed by a one-to-many mapping. This means that a single activity in layer $L{-}1$ can be mapped 
to several activities in layer $L$ by the relation $\vec m'(\vec m)$. One-to-many situations always occur when filters are set to overlap.
The precise form of this relation reads:
\begin{align}
\vec m' (\vec m) &= \left(
\begin{array}{c}
h \Delta + c/\!/(fC') \\
w \Delta + (c/\!/C')\%f\\
c \% C' 
\end{array}\right),\\
\end{align}
where $/\!/$ and $\%$ represent integer and modulo division.

%

\end{document}